\pdfoutput=1

\documentclass[11pt]{article}

\usepackage{ACL2023}

\usepackage{times}
\usepackage{latexsym}

\usepackage[T1]{fontenc}

\usepackage[utf8]{inputenc}

\usepackage{microtype}

\usepackage{array, multirow}
\usepackage{placeins}
\usepackage{xcolor}
\usepackage{soul}
\usepackage{enumitem}
\usepackage{listings}
\usepackage{hyperref}
\usepackage{amsmath}
\usepackage{amssymb}
\usepackage{pifont}
\usepackage{ctable}
\hypersetup{
    colorlinks=true,
    linkcolor=blue,
    urlcolor=blue,
}
\usepackage{graphicx}
\usepackage{svg}
\usepackage{booktabs}
\usepackage{soul}
\usepackage{algorithm2e}

\DontPrintSemicolon
\LinesNumbered
\RestyleAlgo{boxed}
\usepackage[normalem]{ulem}

\SetKwProg{Fn}{function}{:}{}
\SetKwFunction{seqgen}{sequence-generation}
\SetKwFunction{choosenexttriple}{chose-next-triple}

\newlength\mylen

\newcommand{\ie}{i.e.~}
\newcommand{\eg}{e.g.}

\newcommand{\triple}[3]{(\textit{#1}, \textit{#2}, \textit{#3})}

\newcommand{\corpus}{KGConv}

\title{\corpus{}, a Conversational Corpus grounded in Wikidata}
\author{Quentin Brabant\textsuperscript{2}, Gwenole Lecorve\textsuperscript{2}, Lina M. Rojas-Barahona\textsuperscript{2}, Claire Gardent\textsuperscript{1} \\
  \textsuperscript{1}CNRS/LORIA, Nancy, France \\
  \textsuperscript{2}Orange Innovation, Lannion, France \\
  \small\texttt{claire.gardent@loria.fr} \\
  \small\texttt{\{quentin.brabant, gwenole.lecorve, linamaria.rojasbarahona\}@orange.com} \\
}
\date{July 2022}

\begin{document}

\maketitle
\begin{abstract}
    We present \corpus{}, a large, conversational corpus of 71k conversations where each question-answer pair is grounded in a Wikidata fact. 
    Conversations contain on average 8.6 questions and for each Wikidata fact, we provide multiple variants (12 on average) of the corresponding question 
    using templates, human annotations, hand-crafted rules and a question rewriting neural model. 
    We provide baselines for the task of Knowledge-Based, Conversational Question Generation. \corpus{} can further be used for other generation and analysis tasks such as single-turn question generation from Wikidata triples, question rewriting, question answering from conversation or from knowledge graphs and quiz generation. 
\end{abstract}

\section{Introduction}


    
    
    Different from open domain and task-based dialogs, information seeking dialogs are driven by the desire to acquire or evaluate  knowledge. These dialogs are central for instance, in  educational (tutoring a student about a given topic by asking her a set of questions about that topic) and entertainment (quizzes) settings. As large knowledge graphs such as Wikidata\footnote{\url{https://www.wikidata.org/}} have started to emerge, recent years have seen an increasing interest in developing datasets and conversational question answering models that can support such information seeking interactions by grounding dialogs  in factual data.
They often focus on question answering \cite{saha_complex_2018} or provide datasets of restricted size and variety however \cite{christmann_look_2019,lecorve2022sparql2text}.

    In this paper, we focus on information seeking dialogs where as illustrated in Table~\ref{tab:conversation_example}, each question-answer turn is grounded in a single fact. \\ 



Our contribution is two fold.
First, we make available the \corpus{} dataset where, as illustrated
in Table~\ref{tab:conversation_example}, each question-answer pair is
grounded in a Wikidata fact\footnote{We will make our code, data and
pretrained models available online after publication.}. To create a diverse, large scale dataset (70k
conversations), we develop dialogs for eight different topics
(Country, Food, Person, Religion/Ideology, Space Object,
Taxon\footnote{A \textit{taxon} is a population, or group of
populations of biological organisms, e.g. lions or dinosaurs.},
Molecular Entity, and Historical Event). We also include a set of
conversations containing at least one Wikidata property that does not
appear in any of this eight themes to study generalisation. The dataset
include $604$K distinct question-answer pairs covering 458 Wikidata
properties verbalised by 3,879 templates. For each Wikidata triple, there are 12 question variants in average. The dialogs consist of 8.7 question-answer turns in average and for each question in a dialog, there are  three variants: 
(i) an out-of-context form, whose
interpretation is independent of the previous turns, (ii) an
in-context form, where pronouns have been inserted using a rule based
approach and (iii) a synthetic in-context form  obtained by rewriting
in-context forms using a T5 model trained to rewrite questions.

Our second contribution is to establish some baselines for Knowledge-Based, Conversational Question Generation (CQG), the task of generating a question given both a KG fact and a dialog context. 
While much previous work has focused on Knowledge-Based, conversation Question Answering \cite{saha_complex_2018,perez-beltrachini-etal-2023-semantic} or on context-independent, knowledge-based question generation \cite{bordes_large-scale_2015,elsahar-etal-2018-zero,han-etal-2022-generating}, we provide a first investigation of how knowledge-based question generation interacts with conversational context. 
We report results using both automatic metrics and human evaluation. 

    \begin{table}[ht!]
        \newcommand{\interQspace}{0.15cm}
        \center
        \footnotesize
        \begin{tabular}{@{}l@{}r@{}r@{~~}p{6.1cm}}
\toprule
         & \#1 & \textbf{Triple} & \triple{NGC 4833}{part of}{Milky Way}\\[\interQspace]
        & \multirow{6}{*}{\rotatebox[origin=l]{90}{\bf Question variants}}
        &    OOC & \underline{NGC 4833} is part of what astronomical object? \\
        & &  IC & - \\
        & &  SIC & - \\[\interQspace]
        & &  OOC & Where is \underline{NGC 4833} located? \\
        & &  IC & - \\
        & &  SIC & - \\[\interQspace]
        & \multicolumn{2}{r}{\textbf{Answer}} & Milky Way \\[0.10cm]
        \toprule
        & \#2 & \textbf{Triple} & \triple{NGC 4833}{discoverer or inventor}{Nicolas Louis de Lacaille}\\[\interQspace]
        & \multirow{10}{*}{\rotatebox[origin=l]{90}{\bf Question variants}}
        &    OOC & Who was behind the discovery of \underline{NGC 4833}? \\
        & &  IC & - \\
        & &  SIC & Who was behind the discovery? \\[\interQspace]
        & &  OOC & What was the name of the discoverer of \underline{NGC 4833}? \\
        & &  IC & - \\
        & &  SIC & Who discovered this object? \\[\interQspace]
        & &  OOC & Who found \underline{NGC 4833}? \\
        & &  IC & - \\
        & &  SIC & Who found this object? \\[\interQspace]
        & \multicolumn{2}{r}{\textbf{Answer}} & Nicolas Louis de Lacaille \\[0.10cm]
        \toprule
        & \#3 & \textbf{Triple} & \triple{Nicolas Louis de Lacaille}{religion or worldview}{Catholic Church}\\[\interQspace]
        & \multirow{6}{*}{\rotatebox[origin=l]{90}{\bf Question variants}}
        &   OOC & What was \underline{Nicolas Louis de Lacaille}'s religion? \\
        & & IC & What was \underline{his} religion? \\
        & & SIC & - \\[\interQspace]
        & & OOC & What faith did \underline{Nicolas Louis de Lacaille} follow? \\
        & & IC & What faith did \underline{he} follow? \\
        & & SIC & - \\[\interQspace]
        & \multicolumn{2}{r}{\textbf{Answer}} & Catholic Church \\
        \bottomrule
        \end{tabular}
        \caption{Excerpt of a question-answer conversation along with the related triples. The root entity is NGC 4833, from the theme ``space object''. Redundant variants are omitted for readability.}
        \label{tab:conversation_example}
    \end{table}


\section{Related Work}
\label{sec:related_work}

    Question generation from RDF triples is adressed  in~\cite{bordes_large-scale_2015,elsahar-etal-2018-zero,han-etal-2022-generating} and from small KGs depicting multi-hop questions
    \cite{serban2016generating, kumar_difficulty-controllable_2019,bi2020knowledge} and recently in LC-QuAD 2.0 \cite{dubey_lc-quad_2019} and ParaQA \cite{kacupaj_paraqa_2021}. However, these works are limited to the generation of isolated questions, thus no conversational context is under consideration.
    
    CoQA~\cite{reddy_coqa_2019} and QuAC~\cite{choi_quac_2018} are conversational QA corpora, in which answers are extracted from paragraphs, instead of KGs.  
    Similar to \corpus , ConvQuestions~\cite{christmann_look_2019} and CSQA~\cite{saha_complex_2018} are conversational corpora based-on structured knowledge. However, the former does not provide the triples for the questions and contains only $315$ distinct conversations. The latter, despite being a very large dataset covering a wide range of questions types\footnote{Single or multiple triples, entity/numeric/boolean answers, comparative questions}, contains rather unnatural questions with a proprietary formalism which does not directly correspond to Wikidata triples. Recent work proposes to generate questions from SPARQL queries, especially to express complex questions~\cite{lecorve2022sparql2text,perez-beltrachini-etal-2023-semantic}. However, manually annotating questions SPARQL queries is difficult and the authors turn on a small set of $350$ reference questions. In contrast, our dataset focuses on triples and simple questions in order to enable a large set of reference questions, obtained through manually written verbalization templates. 
    
    Our work capitalizes on two existing corpora which provide a correspondence between triples and questions, namely SimpleQuestions \cite{bordes_large-scale_2015} and ZeroShot Relation Extraction \cite{levy_zero-shot_2017}.
    In comparison, the \corpus{} corpus extends the question templates from these corpora by proposing 3,879 new templates, and focusing on 458 properties. Likewise, they focused on isolated questions or short follow-up turns, up to three turns, while \corpus{} contains $8.7$ turns per conversation in average.

    
 
    %

\section{Overview of the Dataset}
\label{sec:overview}

    
In \corpus{}, 
each conversation is focused on a given \emph{root} entity. As illustrated in  Table~\ref{tab:conversation_example}, the first question bears directly on this root entity, while further questions explore new facts about any entity discovered during the conversation (including the root entity itself).
Hence, a conversation can be seen as a small evolving KG, where each turn expands the conversation graph with a new entity and the property which connects it to the existing graph. 

For each root entity, three conversations are derived from Wikidata in order to increase the diversity of the dataset. 
The corpus covers 
eight themes: Country, Food, Person, Religion/Ideology, Space Object, Taxon\footnote{A \textit{taxon} is a population, or group of populations of biological organisms, e.g. lions or dinosaurs.}, Molecular Entity, and Historical Event. The theme of a conversation corresponds to the Wikidata class associated to the root entity  e.g., Person corresponds to the Q215627 class in Wikidata. 
We use Taxon and Space Object as unseen data, data not seen at training time. There is also a set of conversations containing at least one Wikidata property that does not appear in any of the eight regular themes. Table \ref{tab:general_stats} summarizes the size of the dataset for each theme. The number of questions in a conversation is at least 5, at most 19 and 8.6 on average. We provide in total $70,596$ conversations with $603,905$ questions from $63,345$ Wikidata entities and $458$ properties. To enable links with Wikidata and further extensions, Wikidata IDs are provided for all entities and properties along with their natural language labels.

In the conversations, each question has several paraphrases (up to 10, 6.8 on average), and each paraphrase has three versions:
\begin{itemize}
    \item Out-of-Context (OOC):
    this version of a question is produced semi-automatically from a KG fact independent of the conversation context.
    \item In-Context (IC): this version is derived from the OOC version taking the context into account and using rules to substitute repetitions with anaphoric forms.
    \item Synthetic-In-Context (SIC): this version is derived from the IC version
    by applying a generative model trained to rewrite questions. 
\end{itemize}

     \begin{table*}[ht!]
        \centering
        \small
        \begin{tabular}{r|cccc|cccc|cc}
        \toprule
        
        &  entities & properties & triples & conv. & \multicolumn{4}{c|}{number of question-turns} & templates & references \\
       & \multicolumn{4}{c|}{(number in all conversations)}&  train & dev & test & total & \multicolumn{2}{c}{(avg. per question-turn)}  \\ \hline
       person                 &     32k &         327 &    72k &      26k &  185k &  29k &   11k &  226k    & 7.2 & 12.5  \\
        country                &      2k &         171 &     3k &        0.7k &    5k&    0.8k &     0.2k &    6k & 5.5 &  9 \\
        ideology               &      1k &         169 &     2k &        0.4k &    3k &    0.6k &     0.2k &    4k & 6.6 &  11.4 \\
        space object           &      3k &         116 &     6k &       6k &       0 &      0 &   50k &   50k & 7.3 &  12.4 \\
        molecular entity       &     18k &         151 &    38k &      23k &  155k &  25k &    10k &  189k  & 6.4 &  11.7 \\
        historical event       &      5k &         189 &     8k &       5k &   35k &   6k &    2k &   43k  & 5.6 & 10.2 \\
        food                   &      3k &         166 &     4k &       2k &   15k &   2k &    1k &   18k  & 6.1 &  10.4 \\
        taxon                  &      3k &         215 &     5k &       2k &       0 &      0 &   16k &   16k  & 7.9 & 13.6 \\ \hline
        with unseen prop. &     14k &         404 &    24k &       6k &       0 &      0 &   52k &   52k & 6.9 &  12.1 \\ \hline
        \bf{whole dataset}                  &     63k &         458 &   143k &      71k &  398k &  63k &  143k &  604k & 6.8 &  12 \\

        \bottomrule
        \end{tabular}
        \caption{
        Quantitative summary of \corpus{}. Entities, properties and triples can appear in several conversations and several themes but are only counted once. The two last columns show the average number of templates used in a single question-turn, and the number of distinct references (including OOC, IC and SIC versions of all template-based verbalizations).
        }
        \label{tab:general_stats}
    \end{table*}   

\section{Data Collection}
\label{sec:creation_process}

\begin{figure}[h]
        \centering
        \includegraphics[width=0.9\columnwidth]{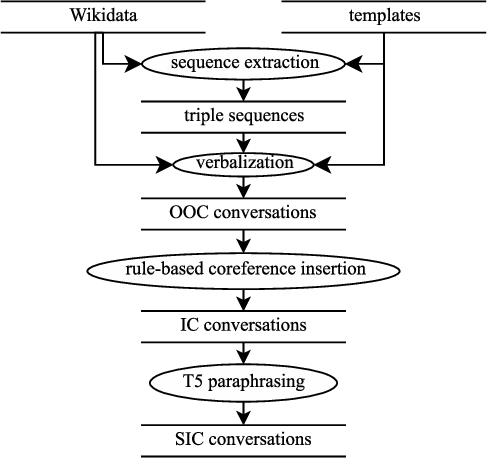}
        \caption{Data flow diagram of the dataset generation.}
        \label{fig:dialgeneration}
    \end{figure}

     The process for creating the corpus is summarized in Figure \ref{fig:dialgeneration} and elicited in the following subsections.
     

    \subsection{Sequence Extraction}
    \label{subsec:seq}
    In this step, we  extract sequences of triples from  Wikidata which will be used to ground the  dialogs.  
    
    We first extract a set of triples from Wikidata such that (i) the subject has at least one English label and the object either has an English label or is a literal (string, number, date, etc.) and (ii) 
    the property belongs to a pre-defined set of Wikidata properties which excludes  ``uninteresting'' properties.  In particular, we excluded properties whose prefix is not ``\texttt{wdt:}'' (to avoid triples that provide meta-information), and properties which link entities to their ID in some other database.
    For any triple $(s,p,o)$ in this set, we then create the reversed triple $(o,-p,s)$ where $-p$ denotes the inverse of  property $p$. In this way,
if our subgraph contains \triple{France}{capital}{Paris}, it also contains  \triple{Paris}{capital}{France} which  permits creating questions about both the subject $s$ and the object $o$ e.g.,  ``What is the capital of France'' and ``Paris is the capital of which country?''. We call the resulting Wikidata subgraph $W$ for World.


        Based on this set of triples, we then create sequences of triples as follows. Each conversation will focus on triples in the neighborhood ${\cal N}(r)$ of a root entity $r$ in $W$.
        This neighborhood is defined as the subgraph of $W$ containing $r$ and all nodes (i.e. entities) that are 1 or 2 edges (i.e. properties) away from $r$; in other words, ${\cal N}(r)$ contains all triples of the form $(r,p,o_1)$ and their successors of the form $(o_1,q,o_2)$.
        Roots were sampled from instances of the 8 themes in Table~\ref{tab:general_stats}, with the condition that their neighborhood is large enough (at least 20 triples)
        to generate 3 reasonably long conversations with enough differences.


        For each root, 3 triple sequences of the form $(t_0, t_1, \dots, t_n)$ were built iteratively by picking triples from ${\cal N}(r)$
        in a greedy stochastic process.
        At each step of the process, the subject of the chosen triple is either the root (\ie $s_i = r$) or an entity, either subject or object, from the previous triple ($s_i = s_{i-1}$ or $s_i = o_{i-1}$). Additionally, a triple cannot appear twice in the sequence.
       The decision to stop or continue the process is made at each time step~$i$ following a probability that increases with $i$: $\Pr_{stop}(i) = 0.06i-0.18$.

\subsection{Verbalization}
\label{sec:template_lib}

Questions are generated using templates like ``What is the capital of \_\_\_?'', where the slot is to be filled by the subject 
of a triple.
Each template~$\tau$ is applicable for a given property $p_\tau$ (\eg ,``capital of''), given required types $\mathcal{S}_\tau$
for the subject slot 
and required types $\mathcal{O}_\tau$
for the object,  
which will be the answer.
This applicability condition on~$\tau$ is denoted as $C(\tau) = (p_\tau, \mathcal{S}_\tau, \mathcal{O}_\tau)$. Then, a triple~$(s, p, o)$ satisfies $C(\tau)$ if: $p_\tau = p$, $\mathcal{S}_\tau \subset \mbox{types}(s)$, and $\mathcal{O}_\tau \subset \mbox{types}(o)$.

    To create a large number of diverse questions for all properties in $W$, 
    we gathered templates from three sources.
    Table \ref{tab:property-question-stats} summarizes the number of templates and their sources. 
    The following sections provide details on the methods used to get templates from each source.

    \subsubsection{Zero-Shot templates}
        The Zero-Shot dataset \cite{levy_zero-shot_2017} contains 1,192 question templates spanning 120 Wikidata properties. Each template $\tau$ was originally created to ask a question for a given property $p$, without taking into account the type of the subject and object.
        Templates grounded on properties that were no longer in Wikidata were discarded.
        For the remained templates, we defined the applicability condition as $C(\tau) = (p, \emptyset, \emptyset)$.
        
    \subsubsection{Simple questions v2 templates}
        We automatically extracted templates from the SimpleQuestions\_v2 dataset \cite{bordes_large-scale_2015}, which contains 108k triple-question pairs, involving 131k distinct entities and 1,837 properties.
        
       As SimpleQuestions\_v2 is based on the Freebase KG, 
        we translated entities and properties into their Wikidata counterpart: we relied both on the Wikidata property P646, that links Wikidata entities to their Freebase counterpart, and on an available mapping between Freebase and Wikidata properties\footnote{\url{https://www.wikidata.org/wiki/Wikidata:WikiProject_Freebase/Mapping}}.
        This allowed us to get Wikidata counterparts for 83,447 entities and 142 properties.
        
       We then extracted templates from question-triple pairs whose triple could be translated to Wikidata. For each such  triple-question pair $((s,p,o),q)$, we created a template $\tau$ by replacing in $q$ the label of $s$ by an empty slot. The applicability condition of $\tau$ was then defined as $C(\tau)=(p,\mbox{types}(s),\mbox{types}(o))$, where $\mbox{types}(x)$ is the set of all types of $x$ in Wikidata.
      Since only a small subset of Freebase entity and triples could be translated into a Wikidata counterpart, many triples were filtered out, so the extracted templates only cover 77 of our properties.

    \subsubsection{New templates}
        We manually created additional templates in three steps: (1) extracting applicability conditions, (2) writing templates corresponding to these conditions, (3) validating written templates.
        
        \paragraph{Step 1: Extracting applicability conditions.}
            From the neighbourhoods of potential roots of all themes, we gathered a set $\{ (s_i, p_i, o_i)\}^N_{i=0}$ of Wikidata triples for which $p_i$ we did not have any template. From this set of triples we generated a set of applicability conditions $\{ (p_i, \mbox{types}(s_i), \mbox{types}(o_i))\}^N_{i=0}$ to be annotated with corresponding templates.
            This resulted in many applicability conditions, with overly specific subject and object types for each property. We solved this problem by iteratively merging conditions with the same property, via semi-automated conceptual agglomerative clustering process.
            Merging two conditions $(p, \mathcal{S}_i, \mathcal{O}_i)$ and $(p,\mathcal{S}_j,\mathcal{O}_j)$ consists in replacing them with a new one $(p,\mathcal{S}_i \cap S_j, \mathcal{O}_i \cap \mathcal{O}_j)$, which necessarily is matched by more triples than the two original ones.
            At the end of this process, applicability conditions that were met by less than 5 triples were discarded.

        \paragraph{Step 2: Template writing.}
            The next step was to write, for each applicability condition $C$, templates that would apply to any triple matching $C$.
           
            
            Three students of an NLP Master program were hired to annotate the question templates. They were native English speakers hired on a short term contract to perform various annotation tasks for NLP. They were paid slightly above the national minimum wage. We provided them with an annotation tool,
            in which one applicability condition at a time was displayed, along with 5 examples of matching triples.
            While annotator were writing templates, their results on these triple were displayed.
            To speed up this process, artificially generated templates were also proposed to the annotators, who could accept or reject them.
            Afterward, accepted artificial templates were treated in the same way as those written by a human annotator.
    
        \paragraph{Step 3: Template validation.}
            To ensure the quality of the resulting templates, all of them were manually filtered by the authors.

    \begin{table}[t]
        \setlength{\tabcolsep}{2pt}
        \centering
        \footnotesize
        \begin{tabular}{lccc}\toprule
        & \bf Properties 
        & 
        \bf  Templates 
        & \bf  Tpl. per prop.
        \\
        & & & min / avg /max. \\
        \midrule
        \bf SimpleQuestions  & ~~77 & ~5,817 & 1 / 76 / 453  \\
        \bf ZeroShot RE  & ~~75  & ~~~~771 & 1 / 10 / ~~31 \\
        \bf New templates & 413 & ~3,879 & 1 / \,~9 / \,~80 \\
        \bf Total & 474 & 10,355 & 1 / 22 / 453  \\
    \bottomrule
        \end{tabular}
        \caption{Statistics on properties and templates for each sources of templates.}
        \label{tab:property-question-stats}
    \end{table}

\subsection{Contextualization}
Applying  the templates from Section~\ref{sec:template_lib} to the triple sequences from Section~\ref{subsec:seq} yields conversations where questions are Out Of Context (OOC) i.e., questions which are independent of the  conversational context. To improve the naturalness of the conversations, we derived 
two in-context versions from these OOC conversations. The first one, \emph{IC}, is obtained by applying hand-crafted rules to introduce coreferences and correct some errors produced during verbalization with templates. The second version (\emph{SIC}, Synthetic-In-Context), results from rewriting IC questions with a T5 model trained to rewrite questions. 

\subsubsection{Post-Processing Dialogs (IC Variants)}  
    \paragraph{Referring Expressions.}
        In Wikidata, an entity can have several labels: one of those is called ``preferred label'' and is meant to be used by default.
        In the OOC version, entities are always referred by their preferred label.
        This step introduces variability by replacing some of the preferred labels with other available labels from Wikidata, according to the following rules:
        (1) the first reference to an entity in the conversation is the preferred label, or contains it as a substring; (2) further mentions are labels that are substrings of the first reference.
        For instance, the entity Q9592 has the preferred label $l_1=$~``Catholic Church'', alternative labels $l_2=$~``Roman Catholic Church'' and $l_3=$~``Roman Apostolic Catholic Church''; if $l_2$ is used as the first reference to the entity, next references will use either $l_1$ or $l_2$ but not $l_3$, since it is not a substring of $l_2$.
        Whenever the subject is a person with a name and surname,  we include its surname in the set of available labels.
    \paragraph{Determiners.}
        Deciding which label should be preceded by ``the'' is not trivial. For example, ``United Kingdom'' and ``Republic of China'' require it, while ``France'' and ``China'' do not. This step handles this problem by asking a BERT language model\footnote{\url{https://huggingface.co/bert-base-uncased}} to fill a mask token inserted before the label; when ``the'' was predicted with a probability at least $0.92$, it was inserted before the label. 
    \paragraph{Tense.}
        We noticed that most templates are written in present tense, while many triples describe facts that are no longer true or concern dead people, past events, etc.
        Questions were rewritten in the past tense\footnote{\url{https://github.com/bendichter/tenseflow}} if the corresponding triple had an ``end time'' qualifier in Wikidata, or if its subject or object was a dead person.

    \paragraph{Rule-based introduction of pronouns.}
        
        Subject mentions are pronominalised  using a rule-based approach: a pronoun is used only if the subject also appears in the triple of the previous question and  if its gender differs from the gender of the object of this triple 
        (to avoid ambiguous pronouns); further rules are used to determine the kind of pronoun to use (for example, if the subject reference is followed by a possessive ``s'', a possessive pronoun should be used, etc.). The pronominalisation rules are given in Appendix \ref{app:pronouns}.

\subsection{Model-based rewritings (SIC Variants)}

To further increase the conversationality degree, a T5-based question rewriting model\footnote{From the base version on HuggingFace: \url{https://huggingface.co/t5-base}.}
was fine-tuned on a training set derived from 2~conversational machine reading QA datasets, namely CANARD~\cite{elgohary-etal-2019-unpack} and CoQAR~\cite{brabant-etal-2022-coqar}.
This training set is made of $142$K instances.
For each instance, the input is a question $q_i$, along with its conversation history $[q_0, a_0, \dots, q_{i-1}, a_{i-1}]$, while the output is a semantically equivalent question whose form is expected to be natural in a conversation, denoted by $q_i^*$.
In some instances, $q_i$ and $q_i^*$ have respectively an OOC and a contextualized form.
In others instances, $q_i$ and $q_i^*$ are equal; these instances correspond to cases where either $q_i$ already has an IC form, or there is no natural way to rewrite it without losing information or bringing ambiguity. Including such cases to the training set enables the model to learn when it should rewrite the input question or not.


At inference time, the 20 best hypotheses are generated by the model for each instance. Then, they are classified into three authorized categories, using a set of expert conditions: (1) coreference with a pronoun (\eg, ``In which country is Kyoto located?'' rewritten as ``In which country is \textit{it} located?''), (2) coreferences with a demonstrative noun phrase (\eg, ``In which country is \textit{this city} located?''), and (3) ellipses (\eg, ``In which country?''). Those that do not belong to any category are filtered out; moreover, to limit possible ambiguities, we prohibit two consecutive reformulations of the same category. Finally, if some hypothesis remain, the one with the highest probability is selected as the rewritten form.

This process was applied on all questions of the IC version, leading to the SIC (Synthetic IC) version. More details can be found in Appendix~\ref{app:sic}.

\begin{table}[]
    \centering
    \footnotesize
    \begin{tabular}{r|cc|c}
    \toprule
    & \multicolumn{2}{c|}{IC} & \multirow{2}{*}{SIC $\neq$ IC}\\
                    & alt. label &  pronoun &  \\ \hline
person                 &         13.8 &     23.7 &       41.7 \\
country                &          4.1 &      8.4 &       47.0 \\
ideology               &          6.4 &     12.3 &       56.7 \\
space object           &          1.6 &     13.7 &       60.4 \\
molecular entity       &         10.8 &      6.0 &       63.7 \\
historical event       &          6.7 &      8.4 &       66.0 \\
food                   &          7.6 &      9.4 &       60.5 \\
taxon                  &          2.0 &     14.2 &       57.2 \\
\hline
unseen properties &         12.6 &     20.2 &       45.5 \\
\bottomrule
    \end{tabular}
    \caption{Percentage of questions using an alternative label and a pronoun in IC questions; percentage of SIC questions that differ from the IC version.}
    \label{tab:my_label}
\end{table}
        


\section{Conversational Question Generation}
\label{sec:tasks}

 Knowledge-Based, Conversational Question Generation extends Question Generation from Knowledge Graph triples \cite{elsahar-etal-2018-zero,han-etal-2022-generating} to a conversational setting:
 instead of generating a question only from a triple, we generate a question from both a triple and the preceding conversational context. 
This raises the additional challenge of generating questions in contextually appropriate forms   e.g., using appropriate referring expressions and ellipses. 
Leveraging the multimodal text/graph nature of our dataset, we explore four ways of representing the context: 
(1) no contextual information at all (Empty), (2) the sequence of previous questions and answers (NL) (3) the sequence of triples underlying the questions and answers (KG) and (4) the sequence of questions and answers with their corresponding triples (NL+KG).



For each of the four variants, we trained a baseline by fine-tuning a T5-small model on the three versions of questions in \corpus{} (OOC, IC and SIC). 
In the train and dev sets, all themes are mixed together.
The number of epochs for training is determined via early stopping.

\section{Automatic Evaluation}\label{sec:auto_eval}

We evaluate each model on the test set using Google-BLEU and BERT-score taking as references all questions associated with the input triples (in OOC, IC, and SIC forms), around 12 references on average.
The results are presented in Table \ref{tab:auto_eval}.

\begin{table*}[h]
\centering
\includegraphics[scale=0.85]{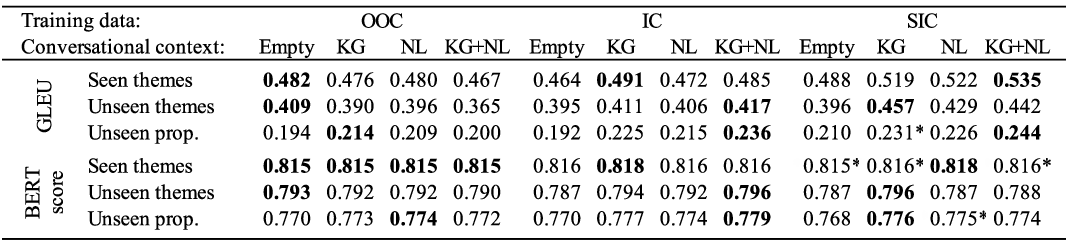}
\caption{\textbf{Results of the automatic evaluation.}
Seen themes are those with a non-empty training set (see Table \ref{tab:general_stats}), unseen themes are space object and taxon.
The scores are obtained by macro-averaging over themes. 
The best score is in bold; lower scores that do not differ significantly from the best one\footnote{$p > 0.05$ in a Mann-Whitney U test} are adorned with (*).}
\label{tab:auto_eval}
\end{table*}

\paragraph{Seen vs. unseen.}
Unsurprisingly, all models obtain lower scores on unseen themes.
Similarly, average scores are lower on unseen properties because the verbalization highly depends on the property of the triple. 

\paragraph{OOC vs. IC vs. SIC training data.}
In term of GLEU, models trained on SIC generally do better than their IC and OOC counterparts. However, this tendency is not confirmed with BERT-scores. Moreover, these better GLEU scores might just be an artifact of the experimental design. Indeed references contain OOC, IC and SIC question versions, and models trained on SIC are the only one to have be trained to generate all three versions. Thus, SIC models might have better GLEU simply because their training data is more in-line with the references.

\paragraph{Conversational context format.}
Adding conversational context to the models trained on IC and SIC questions consistently improves GLEU and BERT-scores. Looking at GLEU scores, it also seems that providing the context in the form of triples (or triples and text) provides better improvement than context in the form of text.
Since conversational context is not required to generate OOC questions, models trained on OOC questions tend to perform better when no conversational context is given, except for unseen properties.  


\section{Human Evaluation}
\label{sec:human_evaluation}

The human evaluation assesses 
both the dataset and the two baselines: IC (KG+NL) and SIC (KG+NL).
We sample conversations from the test set. 
Then, for each of them, we created 4 alternative versions by choosing a specific version of the questions: in two versions, the questions are chosen, respectively, among the IC and the SIC questions of the dataset. The  other two  versions turn on the two IC and SIC baselines used to generate the questions from the triple and the context. 
are fed to an evaluation interface, through which they are rated by human evaluators.
As an illustration, the four versions of a piece of conversation is given in Appendix \ref{app:conv_piece_human_eval}
Table \ref{tab:human_eval_samples} gives the number of conversations rated by evaluators depending on theme and version.
\begin{table}
    \centering
    \footnotesize
    \begin{tabular}{rrcccc}
        \toprule
        & & Ref. & Ref. & Model & Model \\
        & & IC & SIC & IC & SIC \\  \hline
        \multirow{2}{1cm}{(seen)} & person & 10 & 10 & 9 & 8 \\
        & food & 10 & 11 & 10 & 10 \\
        \hline
        \multirow{2}{1cm}{(unseen)} & taxon & 8 & 8 & 8 & 9 \\
        & space object & 6 & 8 & 9 & 8 \\
        \bottomrule
    \end{tabular}
    \caption{Number of rated conversations.}
    \label{tab:human_eval_samples}
\end{table}

\begin{table*}[ht]
    \centering
    \footnotesize
    \setlength{\tabcolsep}{5pt}
    \renewcommand{\arraystretch}{1.2}
    \footnotesize
    \begin{tabular}{cr@{~~}l@{~~~}ccc@{~}c@{~}c@{~}cccc}
        \toprule
        & & & \# of & \# of & \multicolumn{4}{c}{Faithfulness to the triple} & Linguistic &  Semantic & Conversat.\\
        & & & conv. & quest. & \% yes$^\uparrow$ & \% no$^\downarrow$ & \% quite$^\downarrow$ & \% idk & correctness$^\uparrow$ & clearness$^\uparrow$ & naturalness$^\uparrow$ \\
\midrule
    \multirow{4}{*}{\rotatebox{90}{Seen}} & \multirow{2}{*}{Reference from} & IC &         20 &          184 &  \textbf{0.91} &   \textbf{0.05} &  \textbf{0.03} &  0.01 &         4.61 &       4.60 &         \textbf{3.85} \\
&  & SIC &         21 &          183 &  0.77 &   0.15 &  0.08 &  0.01 &         4.64 &       4.09 &         3.52 \\[0.5mm]
& \multirow{2}{*}{Baseline trained on} & IC &         19 &          178 &  0.84 &   0.10 &  0.04 &  0.02 &         4.67 &       \textbf{4.65} &         3.63 \\
&  & SIC &         18 &          161 &  0.80 &   0.12 &  0.07 &  0.01 &         \textbf{4.73} &       4.21 &         3.39 \\
\midrule
\multirow{4}{*}{\rotatebox{90}{Unseen}} & \multirow{2}{*}{Reference from} & IC &         14 &          128 &  \textbf{0.88} &   \textbf{0.06} &  \textbf{0.02} &  0.03 &         4.59 &       \textbf{4.57} &         \textbf{3.93} \\
&  & SIC &         16 &          140 &  0.64 &   0.19 &  0.11 &  0.06 &         4.51 &       3.79 &         3.00 \\[0.5mm]
& \multirow{2}{*}{Baseline trained on} & IC &         17 &          157 &  0.71 &   0.14 &  0.13 &  0.03 &         4.53 &       4.47 &         3.65 \\
&  & SIC &         17 &          153 &  0.76 &   0.10 &  0.12 &  0.02 &         \textbf{4.76} &       4.05 &         2.88 \\
\midrule
\multirow{4}{*}{\rotatebox{90}{All}} & \multirow{2}{*}{Reference from} & IC &         34 &          312 &  \textbf{0.90} &   \textbf{0.06} &  \textbf{0.03} &  0.02 &         4.60 &       \textbf{4.59} &         \textbf{3.88} \\
&  & SIC &         37 &          323 &  0.71 &   0.17 &  0.09 &  0.03 &         4.59 &       3.96 &         3.30 \\[0.5mm]
& \multirow{2}{*}{Baseline trained on} & IC &         36 &          335 &  0.78 &   0.12 &  0.08 &  0.02 &         4.60 &       4.56 &         3.64 \\
&  & SIC &         35 &          314 &  0.78 &   0.11 &  0.10 &  0.02 &         \textbf{4.75} &       4.13 &         3.14 \\
        \bottomrule
    \end{tabular}
    \caption{Scores from human evaluation for conversations about seen, unseen or all themes.}
    \label{tab:human_eval_results}
\end{table*}

\subsection{Evaluation Setup}

The ratings were provided by 15 evaluators from the authors' research center (the authors themselves are not included).
Each evaluator could evaluate up to 50 conversations.
They were told that conversations are automatically generated, but were provided no information about the method employed.
Conversations were presented one by one to the evaluator; for each question-answer pair, the corresponding triple was displayed, and the evaluator had to (1) rate the linguistic correctness of the question on a 5 point scale, (2) evaluate whether the question-answer pair expresses the information of the triple (``yes'', ``quite'', ``no'', ``I don't know'').
In addition, the evaluator had to rate the naturalness of the whole conversation flow.

A second round of evaluation on 61 of the already rated conversations was performed to assess the consistency of ratings among annotators. This evaluation utilizes the IC and SIC versions of the questions (i.e., conversations using baselines were not included in the second round).
This evaluation was performed by 3 annotators who did not participate in the first one.

\subsection{Results}

Results are provided in Table \ref{tab:human_eval_results}. We use 
the Mann-Whitney U and  the $\chi^2$ test to assess significance. The former was used for  correctness, clearness and naturalness scores, since those are evaluated on an ordered scale. The latter was used for  faithfulness scores, since these form a scale that is not completely ordered (because of the ``I don't know'' answer).

\paragraph{IC vs SIC data.}
Comparing the scores of IC and SIC references in the \emph{All} block of Table \ref{tab:human_eval_results}, we see that, while linguistic correctness is roughly the same, SIC references are less clear (clearness 4.59 vs 3.96, p=2e-12), less faithful to the triples (0.90 yes vs 0.71, p=8e-8) and less natural overall (3.88 vs 3.30, p=0.006).

\paragraph{IC vs SIC baseline.}
Baseline (SIC) seems a bit more linguistically correct (4.60 vs 4.75, p=0.001), but baseline (IC) seems clearer (4.56 vs 4.13, p=4e-5) and more natural overall, although it might be due to chance (3.64 vs 3.14, p=0.066). Faithfulness seems to be the same.

\paragraph{Data vs baselines.}
Now let us compare the baselines to the data they were trained on.
For the IC data, the baseline is less faithful to triples (0.90 yes vs 0.78, p=0.00014), otherwise we observe no significant difference.
For the SIC data, the baseline is better on every measure, although we obtain low p-values only for clearness (3.96 vs 4.13, p=0.012) and correctness (4.59 vs 4.75, p=0.0012).

\paragraph{Seen vs. unseen.}
Scores obtained on unseen themes tend to be lower that those obtained on seen themes. Although some of these differences are significant, they happen both for the baseline (IC) and for the SIC references. This suggests that the differences are due to the difficulty of themes rather that the fact that they were seen during the training of baselines or not (more details in Appendix \ref{app:stats}).

\paragraph{Inter-rater agreement.}
We computed Cohen's kappa for each metric (faithfulness, correctness, clearness, naturalness) and obtained, respectively: 0.23, 0,10, 0.22, and 0.14\footnote{
The confusion matrices used for computing kappas can be found in Appendix \ref{app:confused}}.
Low kappas seem due to two factors: (1) the intrinsically subjective nature of the task, which can explain that raters disagree by giving different but close rates, (2) genuine mistakes made by raters (for example, when faithfulness is rated at $yes$ and $no$ by two different raters). It is also possible that differences in raters' fluency had an impact on agreement.
Despite the low agreement, we observed the interesting regularities reported previously in this section.

\section{Conclusion}
We make available \corpus{}, a new conversational dataset grounded in Wikidata where each question-turn in the conversation contains three variants: the out-of-context questions generated from manually annotated  and existing templates; the in-context question automatically generated by rules and the synthetic in-context questions generated by a T5 model.
Although SIC questions have more diverse forms than IC questions, the results of human evaluation suggest that IC questions are more reliable than SIC questions.

We also presented several baselines for the task of question generation and found that generating questions from unseen properties is challenging for these baselines.
An interesting perspective would be to investigate methods for tackling this particular zero-shot task. 

As it provides a large number of references for each question in a dialog, \corpus{} is well suited for other tasks  besides Conversational Question Generation such as in particular, single-turn question generation from facts, 
question rewriting and generation of sequence of question-answer pairs from a Knowledge graph (KG) or vice-versa.

\section{Limitations}
This corpus has been generated semi-automatically, although human annotations were involved in the question templates, the conversations were generated automatically from the KG. As a consequence, in some cases the flow of the conversation may be unnatural, because humans do not usually talk in that way. This might be specially true when conversations involve complex content (e.g. molecular entities, space objects or historical events) that may be difficult to be understood by non experts.

\section{Ethics Statement}
The construction of this corpus involved manual annotation of the question templates. Therefore, we hire three students of an NLP Master program. They were native English speakers hired on a short term contract to perform various annotation tasks for NLP. They were paid slightly above the national minimum wage and they had the right to the social security benefits.

    \bibliographystyle{acl_natbib}
    \bibliography{bib.bib}

\begin{thebibliography}{17}
\expandafter\ifx\csname natexlab\endcsname\relax\def\natexlab#1{#1}\fi

\bibitem[{Bi et~al.(2020)Bi, Cheng, Li, Wang, and Qi}]{bi2020knowledge}
Sheng Bi, Xiya Cheng, Yuan-Fang Li, Yongzhen Wang, and Guilin Qi. 2020.
\newblock Knowledge-enriched, type-constrained and grammar-guided question
  generation over knowledge bases.
\newblock In \emph{Proceedings of the International Conference on Computational
  Linguistics (CICLing)}, pages 2776--2786.

\bibitem[{Bordes et~al.(2015)Bordes, Usunier, Chopra, and
  Weston}]{bordes_large-scale_2015}
Antoine Bordes, Nicolas Usunier, Sumit Chopra, and Jason Weston. 2015.
\newblock \href {https://doi.org/10.48550/arXiv.1506.02075} {Large-scale
  {Simple} {Question} {Answering} with {Memory} {Networks}}.
\newblock ArXiv:1506.02075 [cs].

\bibitem[{Brabant et~al.(2022)Brabant, Lecorv{\'e}, and
  Rojas~Barahona}]{brabant-etal-2022-coqar}
Quentin Brabant, Gw{\'e}nol{\'e} Lecorv{\'e}, and Lina~M. Rojas~Barahona. 2022.
\newblock \href {https://aclanthology.org/2022.lrec-1.13} {{C}o{QAR}: Question
  rewriting on {C}o{QA}}.
\newblock In \emph{Proceedings of the Thirteenth Language Resources and
  Evaluation Conference}, pages 119--126, Marseille, France. European Language
  Resources Association.

\bibitem[{Choi et~al.(2018)Choi, He, Iyyer, Yatskar, Yih, Choi, Liang, and
  Zettlemoyer}]{choi_quac_2018}
E.~Choi, H.~He, M.~Iyyer, M.~Yatskar, W.~Yih, Y.~Choi, P.~Liang, and
  L.~Zettlemoyer. 2018.
\newblock \href {https://doi.org/10.18653/v1/D18-1241} {{QuAC}: {Question}
  {Answering} in {Context}}.
\newblock In \emph{Proceedings of the 2018 {Conference} on {Empirical}
  {Methods} in {Natural} {Language} {Processing}}, pages 2174--2184, Brussels,
  Belgium. Association for Computational Linguistics.

\bibitem[{Christmann et~al.(2019)Christmann, Saha~Roy, Abujabal, Singh, and
  Weikum}]{christmann_look_2019}
Philipp Christmann, Rishiraj Saha~Roy, Abdalghani Abujabal, Jyotsna Singh, and
  Gerhard Weikum. 2019.
\newblock \href {https://doi.org/10.1145/3357384.3358016} {Look before you
  {Hop}: {Conversational} {Question} {Answering} over {Knowledge} {Graphs}
  {Using} {Judicious} {Context} {Expansion}}.
\newblock In \emph{Proceedings of the {ACM} {International} {Conference} on
  {Information} and {Knowledge} {Management}}, pages 729--738. Association for
  Computing Machinery.

\bibitem[{Dubey et~al.(2019)Dubey, Banerjee, Abdelkawi, and
  Lehmann}]{dubey_lc-quad_2019}
Mohnish Dubey, Debayan Banerjee, Abdelrahman Abdelkawi, and Jens Lehmann. 2019.
\newblock \href {https://doi.org/10.1007/978-3-030-30796-7_5} {{LC}-{QuAD} 2.0:
  {A} {Large} {Dataset} for {Complex} {Question} {Answering} over {Wikidata}
  and {DBpedia}}.
\newblock In \emph{Proceedings of the The {Semantic} {Web} (ISWC)}, pages
  69--78. Springer International Publishing.

\bibitem[{Elgohary et~al.(2019)Elgohary, Peskov, and
  Boyd-Graber}]{elgohary-etal-2019-unpack}
Ahmed Elgohary, Denis Peskov, and Jordan Boyd-Graber. 2019.
\newblock \href {https://doi.org/10.18653/v1/D19-1605} {Can you unpack that?
  learning to rewrite questions-in-context}.
\newblock In \emph{Proceedings of the 2019 Conference on Empirical Methods in
  Natural Language Processing and the 9th International Joint Conference on
  Natural Language Processing (EMNLP-IJCNLP)}, pages 5918--5924, Hong Kong,
  China. Association for Computational Linguistics.

\bibitem[{Elsahar et~al.(2018)Elsahar, Gravier, and
  Laforest}]{elsahar-etal-2018-zero}
Hady Elsahar, Christophe Gravier, and Frederique Laforest. 2018.
\newblock \href {https://doi.org/10.18653/v1/N18-1020} {Zero-shot question
  generation from knowledge graphs for unseen predicates and entity types}.
\newblock In \emph{Proceedings of the 2018 Conference of the North {A}merican
  Chapter of the Association for Computational Linguistics: Human Language
  Technologies, Volume 1 (Long Papers)}, pages 218--228, New Orleans,
  Louisiana. Association for Computational Linguistics.

\bibitem[{Han et~al.(2022)Han, Castro~Ferreira, and
  Gardent}]{han-etal-2022-generating}
Kelvin Han, Thiago Castro~Ferreira, and Claire Gardent. 2022.
\newblock \href {https://aclanthology.org/2022.lrec-1.29} {Generating questions
  from {W}ikidata triples}.
\newblock In \emph{Proceedings of the Thirteenth Language Resources and
  Evaluation Conference}, pages 277--290, Marseille, France. European Language
  Resources Association.

\bibitem[{Kacupaj et~al.(2021)Kacupaj, Banerjee, Singh, and
  Lehmann}]{kacupaj_paraqa_2021}
Endri Kacupaj, Barshana Banerjee, Kuldeep Singh, and Jens Lehmann. 2021.
\newblock \href {https://doi.org/10.1007/978-3-030-77385-4_36} {{ParaQA}: {A}
  {Question} {Answering} {Dataset} with {Paraphrase} {Responses} for
  {Single}-{Turn} {Conversation}}.
\newblock In \emph{Proceedings of the The {Semantic} {Web} (ISWC)}, pages
  598--613. Springer International Publishing.

\bibitem[{Kumar et~al.(2019)Kumar, Hua, Ramakrishnan, Qi, Gao, and
  Li}]{kumar_difficulty-controllable_2019}
Vishwajeet Kumar, Yuncheng Hua, Ganesh Ramakrishnan, Guilin Qi, Lianli Gao, and
  Yuan-Fang Li. 2019.
\newblock \href {https://doi.org/10.1007/978-3-030-30793-6_22}
  {Difficulty-{Controllable} {Multi}-hop {Question} {Generation} from
  {Knowledge} {Graphs}}.
\newblock In \emph{Proceedings of The {Semantic} {Web} (ISWC)}, pages 382--398.
  Springer International Publishing.

\bibitem[{Lecorv\'e et~al.(2022)Lecorv\'e, Veyret, Brabant, and
  Rojas-Barahona}]{lecorve2022sparql2text}
Gw\'enol\'e Lecorv\'e, Morgan Veyret, Quentin Brabant, and Lina~M.
  Rojas-Barahona. 2022.
\newblock {SPARQL}-to-text question generation for knowledge-based
  conversational applications.

\bibitem[{Levy et~al.(2017)Levy, Seo, Choi, and
  Zettlemoyer}]{levy_zero-shot_2017}
Omer Levy, Minjoon Seo, Eunsol Choi, and Luke Zettlemoyer. 2017.
\newblock \href {https://doi.org/10.18653/v1/K17-1034} {Zero-{Shot} {Relation}
  {Extraction} via {Reading} {Comprehension}}.
\newblock In \emph{Proceedings of the 21st {Conference} on {Computational}
  {Natural} {Language} {Learning} ({CoNLL} 2017)}, pages 333--342, Vancouver,
  Canada. Association for Computational Linguistics.

\bibitem[{Perez-Beltrachini et~al.(2023)Perez-Beltrachini, Jain, Monti, and
  Lapata}]{perez-beltrachini-etal-2023-semantic}
Laura Perez-Beltrachini, Parag Jain, Emilio Monti, and Mirella Lapata. 2023.
\newblock \href {https://aclanthology.org/2023.eacl-main.184} {Semantic parsing
  for conversational question answering over knowledge graphs}.
\newblock In \emph{Proceedings of the 17th Conference of the European Chapter
  of the Association for Computational Linguistics}, pages 2507--2522,
  Dubrovnik, Croatia. Association for Computational Linguistics.

\bibitem[{Reddy et~al.(2019)Reddy, Chen, and Manning}]{reddy_coqa_2019}
S.~Reddy, D.~Chen, and C.D. Manning. 2019.
\newblock \href {https://doi.org/10.1162/tacl_a_00266} {{CoQA}: {A}
  {Conversational} {Question} {Answering} {Challenge}}.
\newblock \emph{Transactions of the Association for Computational Linguistics},
  7:249--266.

\bibitem[{Saha et~al.(2018)Saha, Pahuja, Khapra, Sankaranarayanan, and
  Chandar}]{saha_complex_2018}
Amrita Saha, Vardaan Pahuja, Mitesh Khapra, Karthik Sankaranarayanan, and
  Sarath Chandar. 2018.
\newblock \href {https://ojs.aaai.org/index.php/AAAI/article/view/11332}
  {Complex {Sequential} {Question} {Answering}: {Towards} {Learning} to
  {Converse} {Over} {Linked} {Question} {Answer} {Pairs} with a {Knowledge}
  {Graph}}.
\newblock \emph{Proceedings of the AAAI Conference on Artificial Intelligence},
  32(1).

\bibitem[{Serban et~al.(2016)Serban, Garc{\'\i}a-Dur{\'a}n, G{\"u}l{\c{c}}ehre,
  Ahn, Chandar, Courville, and Bengio}]{serban2016generating}
Iulian~Vlad Serban, Alberto Garc{\'\i}a-Dur{\'a}n, {\c{C}}aglar
  G{\"u}l{\c{c}}ehre, Sungjin Ahn, Sarath Chandar, Aaron~C Courville, and
  Yoshua Bengio. 2016.
\newblock Generating factoid questions with recurrent neural networks: The 30m
  factoid question-answer corpus.
\newblock In \emph{Proceedings of the Annual Meeting of the Association of
  Computational Linguistics (ACL)}.

\end{thebibliography}

\appendix
\section*{Appendices}
\label{sec:appendix}
\addcontentsline{toc}{section}{Appendices}
\renewcommand{\thesubsection}{\Alph{subsection}.}

\subsection{Pronoun insertion process} \label{app:pronouns}
Here we describe how pronouns where inferred and insert in place of explicit subject mentions.

We consider an OOC question from a conversation, and the triple from which it was generated.
If the subject of the triple is in the previous turn's triple, we try to insert a pronoun with the following steps:
\begin{itemize}
    \item We infer the gender of the subject (male, female, or neutral):
    \begin{itemize}
		\item if the subject has not "human" (Q5) or "fictional character" (Q95074) in his types, then gender is neutral;
		\item otherwise, we use the ``sex or gender'' (P21) wikidata property, if available, to assign male or female to the subject;
		\item if the two previous steps fail, no gender is assigned and the pronoun insertion procedure stops.
    \end{itemize}
	\item A part of speech tagging is done on the question.
	\item We try to insert a pronoun by applying the following rules:
    \begin{itemize}
        \item If the subject's mention is directly preceded by ``a'' or ``of'', no pronoun is inserted;
		\item otherwise, if the subject's mention is directly followed by a possessive marker, that marker is removed and the subject mention is replaced by ``her'', ``his'' or ``its''.
		\item otherwise, if the subject's mention is directly preceded by a proposition or conjunction other than ``of'' and ``that'' or by a non auxiliary verb, the subject mention is replaced by ``her'', ``him'' or ``it'',
		\item otherwise, the subject's mention is replaced by ``he'', ``she'' or ``it''.
    \end{itemize}
\end{itemize}

\begin{table}[t]
    \centering
    \footnotesize
    \begin{tabular}{p{1cm}p{2.2cm}p{2.2cm}c}
    \toprule
       Type & Input  & Output & OK?\\
    \midrule
      \multirow{5}{1cm}{Coref. w/ pronoun} & Which location is Switzerland a component of?   & Which location is it a component of? & $\checkmark$ \\
      \cline{2-4}
      & What was the cause of death of Uriella? & What was her cause of death? & $\checkmark$ \\
      \cline{2-4}
      & What title was held by Martin of Tours? & What title was held by him? & $\checkmark$ \\
      \cline{2-4}
      & Who is in charge of the government of Warsaw? & Who is in charge of the government there? & $\checkmark$ \\
      \cline{2-4}
      & Pierre Chaunel was who's spouse? & Was Chaunel his spouse? & \ding{53} \\
      \midrule
       \multirow{3}{1cm}{Coref. w/ demonstrative noun phrase} & With which country would you associate Gyeonggi Province? & With which country would you associate this province? & $\checkmark$ \\
       \cline{2-4}
      & Which reference work outlined Albigensian Crusade? & Which reference work outlined this conflict? & $\checkmark$ \\
      \cline{2-4}
      & Where are World Council of Churches's headquarters? & Where are these headquarters? & \ding{53}  \\
      \midrule
       \multirow{3}{1cm}{Ellipsis} & What is the public holiday associated with Switzerland? & What is the public holiday? & $\checkmark$ \\
       \cline{2-4}
      & What is the zenith of Eritrea? & What is the zenith? & $\checkmark$ \\
      \cline{2-4}
      & In what geographic region is Eurasia located? & In what geographic region? & $\checkmark$  \\
    \bottomrule
    \end{tabular}
    \caption{Examples of rewritten questions at inference time.}
    \label{tab:app_sic}
\end{table}

\subsection{Question rewritting from SIC}
    \label{app:sic}
    
    This appendix provides details the automatic question rewritting step of the SIC variant for each type of rewritting.
    Below is the details of the experts rules for each transformation type, along with examples in Table~\ref{tab:app_sic}.




\paragraph{Coreference with a pronoun}
(\eg, ``\textit{In which country is Kyoto located?}'' rewritten as ``\textit{In which country is \textit{it} located?}'')
A candidate rewritten question is considered an acceptable if it contains one of the following words: ``he'', ``it'', ``she'', ``they'', ``his'', ``its'', ``her'', ``their'', ``them'', ``hers'', ``theirs'', ``there''. I should also be close enough from the original question. This is handled by removing the entities from previous turns from both the original and candidate questions, and then aligning the two word sequenecs using a Levenshtein distance. If only one word is different or less than $25\%$ of the candidate question words are different, the candidate question is validated.

\paragraph{Coreference with a demonstrative noun phrase}
(\eg, ``\textit{In which country is \textit{this city} located?}'')
A candidate question is considered as acceptable if it contains one of the article ``this'' or ``these'' followed by a word. To avoid ill-formed questions like ``\textit{which region is this region in?}" or "\textit{who is this located in?}'', an additional constraint makes it sure that the word after the demonstrative pronoun is not duplicated in the question, nor too short.

\paragraph{Ellipsis}
(\eg, ``\textit{In which country?}'')
A candidate rewritten question is considered as an ellipsis if (1) it is a prefix of the original question to make sure the target of the questions remains the same, and (2) it does not include any entity from the previous question and answer turns.

Only qualitative evaluations were performed but, as illustrated in Table~\ref{tab:app_sic}, the rewritten questions are OK in a large majority of cases.

\subsection{Example of conversation}\label{app:conv_piece_human_eval}

\begin{table}[h!]
    \centering
    \includegraphics[width=0.48\textwidth]{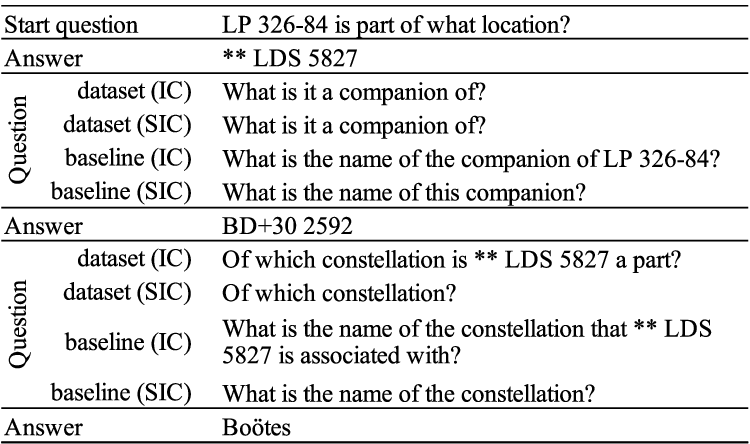}
    \caption{Piece of data from the human evaluation. The four versions of one conversation are shown together in the table.}
\end{table}

\subsection{Statistical analyses of human evaluation results}
    \label{app:stats}
    
    Table \ref{tab:pvalues_seen_unseen} provides the p-values resulting from statistical tests aiming to assess whether the differences in scores between seen (person and food) and unseen themes (taxon and space) was due to chance (scores tend to be higher in the seen category). Several p-values are low, which tends to show that unseen themes are indeed more difficult. However, it does not show that they are more difficult because they were not seen during training. Indeed, the distinction between seen and unseen is only relevant for baselines; yet, SIC questions from the dataset obtain significantly lower scores on unseen themes.
    It is therefore reasonable to assume that the difference between seen and unseen theme is due to the fact that some themes are intrinsically harder to handle for a pretrained T5.
    \begin{table}[]
        \centering
        \footnotesize
        \setlength{\tabcolsep}{2pt}
        \begin{tabular}{lrcccc}
        \toprule
           &                & clearness & naturalness & correctness & faithfulness \\ \hline
        Reference &        IC        & 0.103  &   0.262   & 0.172    & 0.321 \\
        \cline{2-6}
        from      &  SIC       & 0.013  &  0.062   & 0.091   & 0.007 \\
        \hline
        Baseline   & IC     & 0.054 &      0.493 &   0.028 &   0.01 \\
        \cline{2-6}
        trained on & SIC     & 0.221 &   0.09 &    0.328 &     0.539 \\ \bottomrule
        \end{tabular}
        \caption{$p$-values of seen/unseen differences:}
        \label{tab:pvalues_seen_unseen}
    \end{table}

\subsection{Inter-rater agreement}\label{app:confused}
    Computing the Cohen's kappa of ratings from Table \ref{tab:human_eval_results} gives poor scores.
    However, the confusion matrices suggest that, although the exact rate given to a question has a high degree of subjectivity,
    raters tend to give close ratings.
    \begin{table}[h]
        \centering
        \includegraphics[width=0.95\columnwidth]{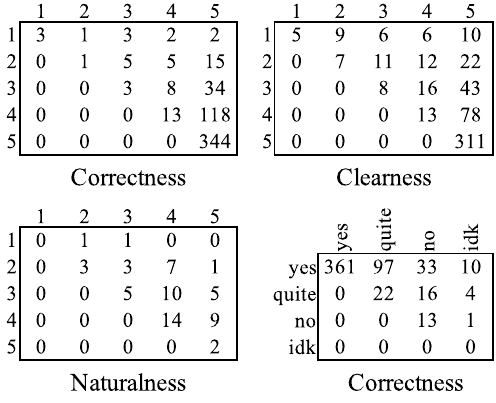}
        \caption{Confusion matrices of human ratings.}
        \label{tab:confusion}
    \end{table}

\end{document}